\documentclass[sigconf]{acmart}

\usepackage{booktabs} 
\usepackage{tabu}
\usepackage[inline, shortlabels]{enumitem}
\usepackage{algorithm,algorithmic}
\usepackage{subfig}
\newcommand{\eat}[1]{}

\copyrightyear{2018} 
\acmYear{2018} 
\setcopyright{acmcopyright}
\acmConference[CIKM '18]{2018 ACM Conference on Information and Knowledge Management}{October 22--26, 2018}{Torino, Italy}
\acmBooktitle{2018 ACM Conference on Information and Knowledge Management (CIKM'18), October 22--26, 2018, Torino, Italy}
\acmPrice{15.00}
\acmDOI{10.1145/XXXXXX.XXXXXX}
\acmISBN{978-1-4503-6014-2/18/10} 
\editor{Jennifer B. Sartor}
\editor{Theo D'Hondt}
\editor{Wolfgang De Meuter}

\hypersetup{draft} 
\begin{document}
\title{Coarse-to-Fine Annotation Enrichment for Semantic Segmentation Learning}
\eat{
\titlenote{}
\subtitle{}
\subtitlenote{}}


\author{Yadan Luo}
\authornote{Work done in part during the internship at DJI.}
\affiliation{%
	\department{School of Information Technology and Electrical Engineering}
	\institution{The University of Queensland}
}
\email{lyadanluol@gmail.com}

\author{Ziwei Wang}
\affiliation{%
	\department{School of Information Technology and Electrical Engineering}
	\institution{The University of Queensland}
}
\email{ziwei.wang@uq.edu.au}

\author{Zi Huang}
\affiliation{%
	\department{School of Information Technology and Electrical Engineering}
	\institution{The University of Queensland}
}
\email{huang@itee.uq.edu.au}

\author{Yang Yang}
\affiliation{%
	\department{Center for Future Media and School of Computer Science and Engineering}
	\institution{University of Electronic Science and Technology of China}
}
\email{dlyyang@gmail.com}

\author{Cong Zhao}
\affiliation{%
	\department{}
	\institution{DJI Innovation Inc}
}
\email{cong.zhao@dji.com}

\renewcommand{\shortauthors}{Luo, Y. et al.}

\begin{abstract}
	Rich high-quality annotated data is critical for semantic segmentation learning, yet acquiring dense and pixel-wise ground-truth is both labor- and time-consuming. Coarse annotations (\textit{e.g.,} scribbles, coarse polygons) offer an economical alternative, with which training phase could hardly generate satisfactory performance unfortunately. In order to generate high-quality annotated data with a low time cost for accurate segmentation, in this paper, we propose a novel annotation enrichment strategy, which expands existing coarse annotations of training data to a finer scale. Extensive experiments on the Cityscapes and PASCAL VOC 2012 benchmarks have shown that the neural networks trained with the enriched annotations from our framework yield a significant improvement over that trained with the original coarse labels. It is highly competitive to the performance obtained by using human annotated dense annotations. The proposed method also outperforms among other state-of-the-art weakly-supervised segmentation methods.
\end{abstract}

%



\maketitle

\section{Introduction}
\begin{figure}[!htb]
	\centering
	\includegraphics[width=.9\linewidth]{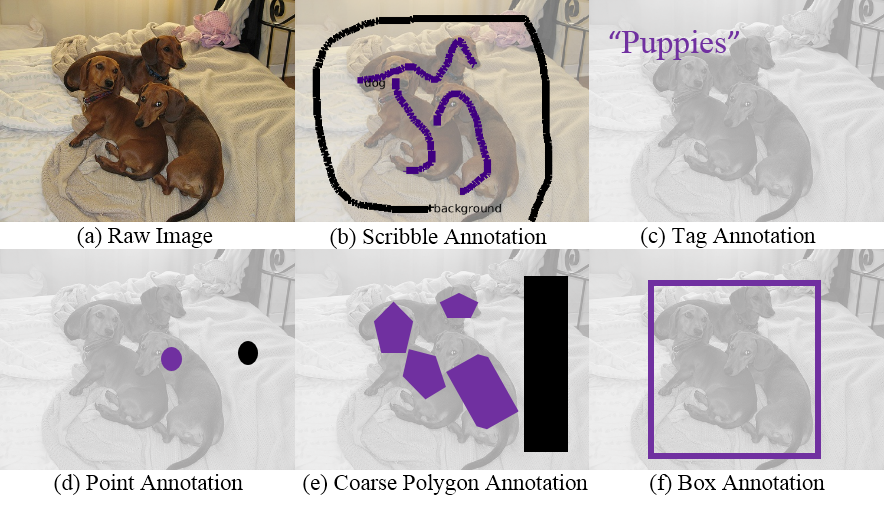}\\
	\caption{Examples of the different forms of coarse annotations for semantic segmentation. }
	\label{fig:example}
\end{figure}
Fully-supervised semantic segmentation has recently achieved prominent progress~\cite{DBLP:conf/cvpr/BertasiusTYS17,DBLP:conf/cvpr/LiLLLT17,fcn,deeplab,pspnet} while triggering a huge demand of dense annotated data. The conventional protocol~\cite{cityscapes,pascalvoc} requires annotators to click along region boundaries until generating a closed polygon and eventually produce a pixel-level label map. Obviously, obtaining large corpus of accurate label maps requires expensive and tedious human labor, especially when confronted with objects that are not suitable for polygon representations (\textit{e.g.,} round sign board, sparse branches). Besides, from a cognitive perspective~\cite{extreme}, humans are not good at putting clicks on narrow boundaries, resulting in incorrect labelling. As report in ~\cite{cityscapes}, annotating a single 1024p image from Cityscapes dataset with quality control averagely costs 1.5 hours.

\begin{figure*}[!htb]
	\centering
	\includegraphics[width=1\linewidth]{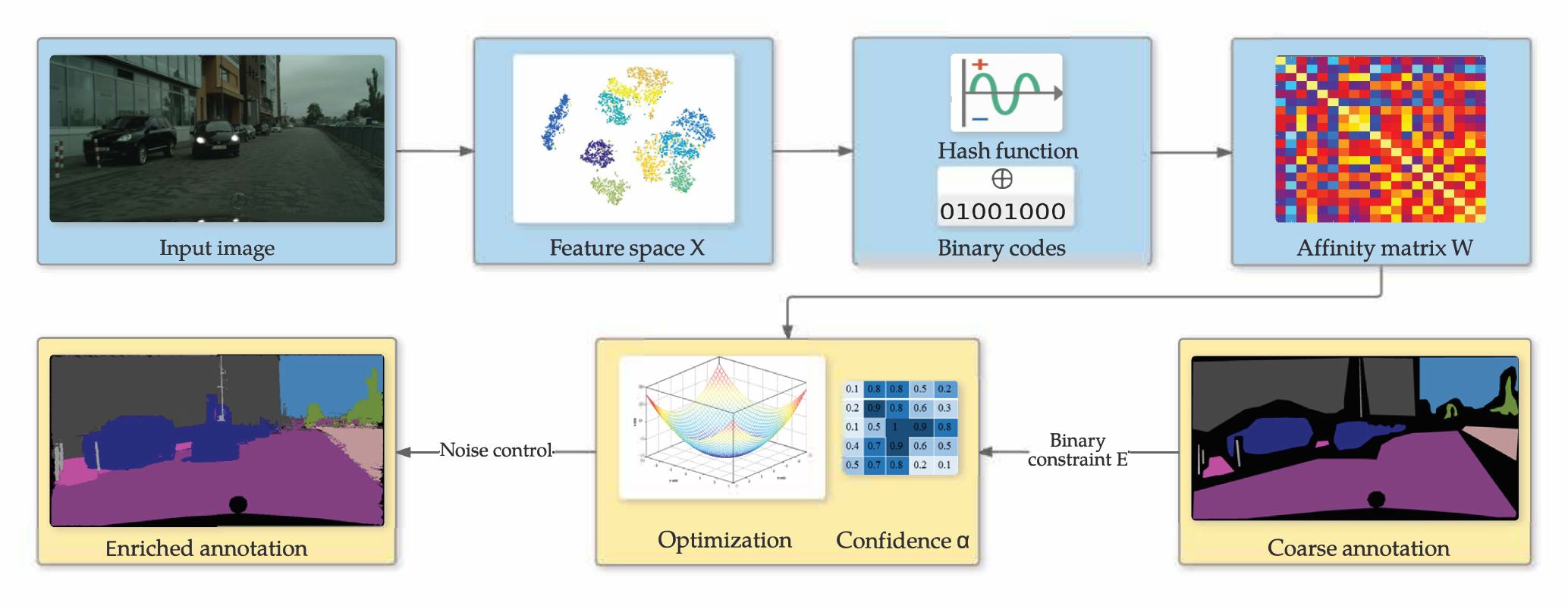}
	\caption{The proposed framework. The blue blocks show the offline components and the yellow ones present online processing. Each time when coarse annotations come along, markups are viewed as binary constraint that helps to update the prediction of unlabeled pixels.
	}%
	\label{fig:pipeline}
\end{figure*}

To make annotation more efficiently, \cite{polyrnn,polyrnn1} propose a semi-automatic CNN-RNN framework for sequentially predicting vertices of the polygon outlining the objects inside the given bounding box. It benefits human annotators via relieving their work load. However, it still suffers from the limitation of polygon representation and over-depends on the quality of box.

Could we find an alternative to dense annotations? As \cite{point,cityscapes} report, annotating coarsely not only allows annotators to easily draw on the region where they feel confident, but also accelerates 7 - 13$\times$ faster than polygon drawing. Coarse annotations provide various forms of user interactions to boost annotating efficiency, where some examples are shown in Figure \ref{fig:example}. However, directly applying those coarse annotations is hard to generate competitive results due to its sparse supervision. 

Facing a dilemma either to consume large amount of labor and time for dense supervision or carefully derive weakly-supervised models for propagating coarse annotations, in this paper, we open a new direction of coarse-to-fine annotation enrichment for generating sufficient reliable labels from coarse annotations with a low time cost. A generalized efficient framework is proposed to automatically generate dense annotations based on the original coarse annotations of images, where fine annotations are achieved with a negligible time cost. As Figure~\ref{fig:pipeline} shows, our framework firstly constructs the feature space and the affinity matrix on the raw image to preserve data structure. Then it propagates sparse user markups to update the beliefs of the classifier for unlabeled points. With the help of the nonlocal principle, we relax the binary constraint and formulate this framework as a convex problem with a closed-form solution. The proposed method could be even generalized for refining sparser and more challenging coarse annotations such as scribbles.

Extensive experiments have been conducted on two public benchmark datasets, \textit{i.e.}, the Cityscapes and PASCAL VOC 2012, with various configurations to evaluate the effectiveness of the proposed methods from different perspectives. The experimental results have evidenced that our annotation enrichment framework can be a cost-effective and flexible solution for semantic segmentation learning.

Our \textbf{main contributions} are summarized as follows:
\begin{itemize}
	\item Instead of heavily relying on dense labelling or sophisticated weakly-supervised models for segmentation, we open a new direction of improving segmentation performance by proposing a coarse-to-fine annotation enrichment regime to enrich the labeled training data efficiently.
	\item By incorporating basic properties (\textit{e.g.,} color, spacial location) and regional patterns of pixels, our feature space balances both sharpness and smoothness, which enables the expanded mask to acquire more accurate contours.
	\item Unsupervised hashing is utilized to deal with high-dimensional pixel affinity with bit-wise xor operations for query processing, which significantly reduces the time cost without compromising on performance.
	\item In light of the nonlocal principle, our loss function with binary markups constraint is relaxed to be differentiable and has a closed-form solution.
\end{itemize}
The rest of the paper is organized as follows. In Section 2, related work including weakly-supervised segmentation and annotation refinement is presented. The problem definition and details of our proposed framework are provided in Section 3, followed by comprehensive performance studies and a conclusion in Section 4 and 5 respectively.

\section{Related Work}
\subsection{Annotation Refinement}
There is rare literature exploring refinement of coarse annotations. In \cite{objectcontour}, a dense conditional random field (CRF)~\cite{densecrf} is used to refine imperfect annotations, which positively guide encoder-decoder (CEDE) structure to learn a better model. This case potentially proves that post-processing methods could also be used for annotation refinement. CRF is widely applied in existing semantic segmentation~\cite{deeplab,crfasrnn,DBLP:conf/cvpr/ShenGYZ17,DBLP:conf/eccv/ChandraK16,DBLP:journals/tip/ZandDHM16} and many experiments have shown its validity and scalability. However, its high computation load and sensitivity to parameter selection attenuate its practicability. From another prospective, dense CRF is not specially designed for refining annotations but for general segments. Treating every pixel equally degenerates its capacity of preserving original labeled area from coarse annotations that of high confidence.
\subsection{Weakly-supervised Segmentation}
Due to the lack of fine annotated data, great efforts have been placed in directly applying various forms of weak annotations into training segmentation models, which are named as weakly-supervised segmentation. According to the coarse annotations they are trained with, the weakly-supervised segmentation models can be roughly classified into four categories including \textit{image-tag based~\cite{image,zhang2013sparse,AE-PSL,AF-SS}, bounding box based~\cite{wssl,boxsup}, point based~\cite{point}, and scribble based~\cite{scribbleup,lazysnapping,rother2004grabcut}. }

The image-tag based segmentation takes into account the image-level tags to decide the presence or absence of classes. To learn the segmentation model from image labels, \cite{image} casts each image as a bag of pixel-level-instances and define a pixel-wise, multi-class adaptation of Multi-instance Learning (MIL) for the loss. The bounding box based methods usually exploit easier-to-obtain detection result with human verification to generate pixel-wise labels inside with
GrabCut\cite{rother2004grabcut} tool or its modified versions. Point based approaches~\cite{point} seem more economical, which combine objectness prior and one point per class to train a CNN for segmentation.
Scribble based methodologies require annotators to provide a few easy-drawing scribbles on the regions which they feel confident with. \cite{scribbleup} derives a graphical model by iteratively propagating information from scribbles and learns network parameters. However, due in part to sparsity of coarse supervision, in most cases, the performance yielded by weakly-supervised segmentation are not competitive to supervised methods.
\section{Annotation Enrichment}
\subsection{Problem Definition}
Suppose that we have a raw image $I$ consisting of $n=m\times l$ pixels, where $m$ and $l$ are the dimensions of $I$. Annotations of a total number of $c$ classes in $I$ are represented as a set of binary maps $S = \{s_{1}, s_2, ..., s_{c}\}$, where $s_k \subset \{0,1\}^n$, $k=1..c$, indicating the presence of each pixel annotated to the $k$-th class. Given a set of coarse annotations of $I$, we aim to enrich them to a finer scale by expanding the original coarse annotation masks. The objective belief masks are denoted as $A = \{\alpha_1, ..., \alpha_c\}$, where $\alpha_k \subset (0,1)^n$, $k=1..c$. $\alpha_{k_i}$ indicates the confidence of the $i$-th pixel that belongs to the $k$-th class.
\subsection{Loss Function}

In each expanding step of generating $\alpha_k$ based on $s_k$, we define the loss function as follows to capture data connectivity and discrimination. $S_{target}$ and $S_{others}$ denote the indices of the pixels with the value of 1 in $s_k$ and the rest markups respectively.

\begin{equation}
\label{eq:1}
\begin{split}
\min_{\alpha_k}L(\alpha_k) &= \frac{1}{2}\sum_{i,j=1}^nw_{i,j}(\alpha_{k_i} - \alpha_{k_j})^2\\&
+ \lambda \sum_{i\in S_{others}}\alpha_{k_i}^2 + \lambda\sum_{i\in S_{target}}(\alpha_{k_i}-1)^2,
\end{split}
\end{equation}

where $w_{i,j}$ is an affinity weight between $i$-th pixel and $j$-th pixel given by a kernel function and $\lambda$ is a penalty parameter. The first term serves as a relationship constraint, that is, points of close affinity are more likely to share a similar confidence score of belonging to the $k$-th class. Moreover, the learned framework is also expected to discriminate the distribution of inhomogeneous pixels so that the rest terms are added. So far, the presented loss function is NP hard due to its binary constraint. Therefore, in the following section, we introduce the nonlocal principle to relax the problem and give a closed-form solution in Section 3.6.

\subsection{Nonlocal Principle}
With the inspiration of~\cite{nonlocalmatting}, our method also capitalizes on the nonlocal principle~\cite{nonlocal} in constructing affinities so that good graph clusters are obtained for respective class.

The working assumption of the nonlocal principle is that the $i$-th denoised pixel is a weighted sum of the pixels of similar appearance with the weights given by a kernel function $w(i,j)$:
\begin{equation}
\begin{split}
E[X(i)]~&\approx\sum_{j} X(j)w(i,j)\frac{1}{D_{i}},\\
w(i,j)~&=~exp(-\frac{1}{h_1^2}\left\|X(i) - X(j)\right\|_g^2 - \frac{1}{h_2^2}d_{ij}^2),\\
D_i~&=~\sum_{j}w(i,j),
\end{split}
\label{eq:3}
\end{equation}
where $X(i)$ is a feature vector computed using the information around $i$-th pixel, $d_{ij}$ is the distance between the $i$-th and $j$-th pixel, $\left\|\cdot\right\|_g$ is a norm weighted by a center-weighted Gaussian, and $h_1$ and $h_2$ are constants found empirically.

As $\alpha_{k_i}$ measures the possibility of the $i$-th pixel belonging to the $k$-th class, pixels of the same class are expected to share similar $\alpha$ values. By analogy of equation \eqref{eq:3}:
\begin{equation}
\begin{split}
E[\alpha_{k_i}]~&\approx\sum_{j} \alpha_{k_j} w(i,j)\frac{1}{D_{i}},\\
\Rightarrow D_i\alpha_{k_i}&\approx w(i,\cdot)^T\alpha_k.
\end{split}
\label{eq:4}
\end{equation}

We could infer from \eqref{eq:4}:
\begin{equation}
\begin{split}
\alpha_k^T L_c \alpha_k \approx 0,\\
\end{split}
\label{eq:5}
\end{equation}
where $L_c = (D-W)^T(D-W)$ is called clustering Laplacian. This basically solves the quadratic minimization problem, the first term of \eqref{eq:1}. For fast implementation, we replace $L_c$ with the Laplacian $L = D - W$ due to its sparsity, which enables 100 times faster without yielding compromising results. Without loss of generality $L$ is used in the rest section.

When dealing with multi-class ($c >1$) coarse annotations, it is also expected to satisfy the summation property, that is, the estimated $\alpha_k$ at any given pixel should sum up to 1:
\begin{equation}
\begin{split}
\sum_{k=1}^c\alpha_{k_i} &=\textbf{1}, ~~ \forall i = 1..n\\
(L + \beta E)\textbf{1} &= \beta E\textbf{1} = \beta s.
\end{split}
\label{eq:6}
\end{equation}
where $E = diag(s)$, and $\beta$ is a constant controlling user's confidence on the markups. The nullspace of Laplacian $L$ is \textbf{1} a constant vector with all 1's. Thus we could achieve,
\vspace{-0.2cm}
\begin{equation}
\begin{split}
(L + \beta E)\sum_{k=1}^c\alpha_k &= \beta s.
\end{split}
\label{eq:7}
\end{equation}

\subsection{Computing Affinity Matrix $W$}

In this section, we describe how to build an affinity matrix $W\in R^{n\times n}$. Computing the affinity matrix $W$ involves collecting nonlocal neighborhoods before their feature vectors $X(\cdot)$ are matched using $k(i,j)$. Traditionally, k-d tree~\cite{kdtree} harnesses tree properties to quickly eliminate large portions of the search space, where each node could easily find its homogeneous pixels. However, resulting from `curse of dimensionality', its performance will sharply degrade even no better than exhaustive search~\cite{DBLP:reference/cg/2004} as the number of dimensions increases.

To maintain affinity quality, we convert this problem to approximate nearest neighbor search~(ANN) problem that has been widely investigated~\cite{SDH,robusthash,zeroshothash,asym}. By incorporating classical unsupervised hashing method ITQ~\cite{gong2013iterative}, the algorithm explores the data local structure that helps to learn a hash function, thus enabling projection $\hat{W}$ from original features $X$ to $\tau$-length binary codes $B\in \{-1, 1\}^{n\times\tau}$ as the following equation shows:
\begin{equation}
B = sgn(X\hat{W}),
\end{equation}
where $sgn(\cdot)$ denotes a signum function that extracts the sign of a real number, $X\in\mathbb{R}^{n\times z}$,  $\hat{W}\in\mathbb{R}^{z\times\tau}$ is the learned transformation, and $z$ denotes the feature dimension of $X$. In this way, we could measure the similarity between pixels by computing the Hamming distances between binary codes. When computing the $i$-th row of affinity matrix, we retrieve top $K$ nearest neighbors of the $i$-th pixel~($K$ varies from 5 to 30) with the minimum Hamming distance, where only efficient bit-wise xor operations are required.\\
\begin{figure}[!htb]
	\centering
	\includegraphics[width=.9\linewidth]{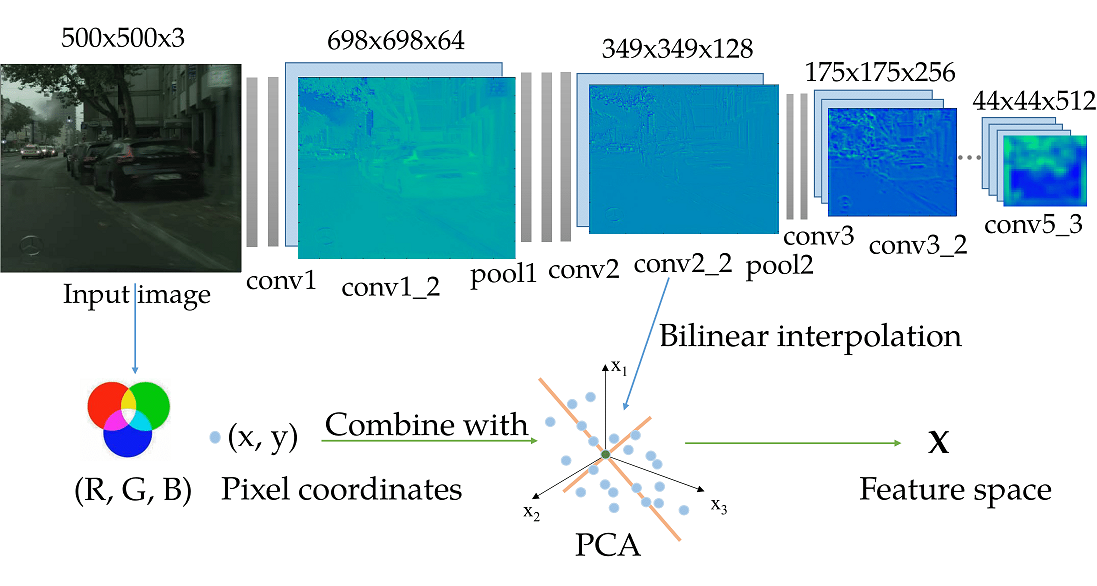}\\
	\caption{The illustration of computing feature vector. By incorporating local features (\textit{e.g.,} color, spacial location) and regional patterns of pixels, our feature space balances both sharpness and smoothness, which enables the expanded mask to acquire more accurate contours. }
	\label{fig:feature}
\end{figure}

\subsection{Computing Feature Vector $X$}

Essentially, each pixel can be represented by its $(r, g, b)$ color and the coordinates in $I$. Could we import other features which has strong discriminative power of examining shapes/boundaries of semantic segments? In light of the development of neural networks, we extract embedded features from conv2 layer of pretrained FCN-8s and do bilinear interpolation to image-scale as the Figure \ref{fig:feature} shows. Note our method is agnostic to the choice of pretrained models, since shallow filters are usually sensitive to edges, texture and color. Assigning those features to every pixel makes the original feature space smoother and more sensitive to local structure. However, a large number of feature map channels may potentially suppress the positive impact of RGB information and spacial correlation, by which sharper guidance is provided. To leverage different features in a balanced manner, we apply principal component analysis on the extracted features maps to reduce the dimensionality. Therefore, the feature of the $i$-th pixel of $I$ that cooperates with spacial coherence, appearance and patterns could be formulated as:
\begin{equation}
\begin{split}
X(i) = (R, G, B, F_i, x, y),
\end{split}
\end{equation}
where $F$ indicates dimension-reduced feature maps. For efficient implementation, we utilize singular value decomposition for principal component transformation. To further investigate the validity of fused features, we conduct the ablation study on feature selection and visualize the outcome in the later section.

\subsection{Closed-form Solution}

By considering the constraint \eqref{eq:6} from section 3.3, a complete form of equation \eqref{eq:1} is rewritten as:

\begin{equation}
\begin{split}
\min_{\alpha_{k}}~L(\alpha_{k})=&\alpha_{k}^TL\alpha_{k} + \lambda\sum_{i\in S_{others}}\alpha_{k_i}^2 + \lambda\sum_{i\in S_{target}}(\alpha_{k_i}-1)^2,\\
=& \alpha_{k}^TL\alpha_{k} + \lambda \sum_{i\in s} \alpha_{k_i}^2 + 2\lambda S_{target}^T\alpha_{k} + \lambda|S_{target}|,\\
=& \alpha_{k}^T(L+\gamma E)\alpha_{k} - 2\lambda S_{target}^T\alpha_{k} + \lambda|S_{target}|,
\end{split}
\end{equation}
\label{eq:8}
where $\lambda|S_{target}|$ is a constant found empirically,  and $\gamma = \beta / \lambda$. It has closed-form solution:
\begin{equation}
\begin{split}
\frac{\partial L(\alpha)}{\partial\alpha} = 2(L +\gamma E)\alpha - 2\lambda S_{target} = 0,\\
\alpha = (L - \gamma E)^{-1}(\lambda S_{target}).
\end{split}
\label{eq:8}
\end{equation}

\begin{table*}[!htb]
	\centering
	\makebox[0pt][c]{\parbox{1\textwidth}{%
			\begin{minipage}[b]{1\hsize}\centering
				\caption{Comparisons of our method and dense-CRF with coarse annotated data on the Cityscapes datasets \textit{w.r.t} IoU and Acc.}
				\begin{tabu}to 1\textwidth{X[2.5,c]|X[c]X[1.3,c]X[c]X[c]X[1.1,c]X[c]X[c]X[c]X[c]X[1.3,c]X[1,c]X[1.2c]X[c]X[c]X[c]X[c]X[c]X[1.3,c]X[1.2,c]|X[1.5,c]}
					\hline
					Method & road & swalk & build & wall & fence & pole & tlight & sign & veg. & ter. & sky & pers. & rider & car & truck & bus & train & mbike & bike & Acc.\\
					\hline
					coarse &44.4&61.4&34.3&42.8&29.8&32.9& 45.7&57.8&30.3&62.4&47.3&37.3&66.0&49.4&63.1&61.3&40.1&37.6&23.9&60.9\\
					CRF &59.7&77.8&39.7&44.4&27.2&33.5& 44.2&76.9&44.6&85.0&53.4&42.8&77.1&58.0&68.8&71.8&51.7&46.2&11.2&70.4\\
					Fil.(0.7) &58.3&76.1&40.3&\textbf{47.6}&34.2&41.1& 55.4&76.9&43.2&84.5&\textbf{60.6}&49.3&\textbf{80.6}&58.4&\textbf{72.0}&\textbf{74.0}&56.9&\textbf{50.4}&38.9&76.9\\
					Fil.(0.3)  &\textbf{64.7}&\textbf{79.6}&\textbf{43.4}&44.9&\textbf{35.9}&\textbf{42.8}& \textbf{57.8}&\textbf{80.0}&\textbf{49.0}&\textbf{89.9}&59.8&\textbf{51.2}&77.7&\textbf{62.1}&68.3&68.7&\textbf{60.2}&48.7&\textbf{54.4}&\textbf{82.1}\\
					\hline
				\end{tabu}
				\label{crf_cityscapes}
			\end{minipage}
	}}
\end{table*}

\begin{table*}[!htb]
	\centering
	\makebox[0pt][c]{\parbox{1\textwidth}{%
			\begin{minipage}[b]{1\hsize}\centering
				\caption{Comparisons of our method and dense-CRF with coarse annotated data on the PASCAL VOC 2012 datasets \textit{w.r.t} IoU and Acc.}
				
				\begin{tabu}to 1\textwidth{X[3,c]|X[c]X[1.3,c]X[c]X[c]X[1.1,c]X[c]X[c]X[c]X[c]X[1.3,c]X[1,c]X[1.2c]X[c]X[c]X[c]X[c]X[c]X[c]X[1.3,c]X[1,c]X[1.3,c]|X[1.3,c]}
					\hline
					Method & bg & acr & bic& bir & boat &bot & bus & car & cat & chair & cow & din & dog & hor & mot & pcr & pot & shc & sof & tra & tv &Acc.\\
					\hline
					coarse &60.3&34.3&17.6&26.1&31.2&27.6& 19.3&22.8&20.0&27.5&25.2&22.3&21.3&26.2&22.4&25.8&24.2&25.6&23.4&20.8&2.7&62.5\\
					CRF &83.1&48.1&22.0&51.6&46.0&58.5& 68.6&61.1&74.1&47.3&63.9&75.8&68.5&62.1&58.9&61.9&47.2&61.3&74.0&67.9&9.8&83.6\\
					Fil.(0.3) &86.6&55.2&29.3&68.2&\textbf{57.1}&58.2& 74.3&59.6&78.3&\textbf{55.7}&73.9&75.5&71.5&65.4&58.4&66.3&47.2&\textbf{69.1}&\textbf{75.4}&\textbf{71.2}&10.3&87.0\\
					Fil.(0.7)  &\textbf{87.0}&\textbf{58.8}&\textbf{33.2}&\textbf{76.6}&51.9&\textbf{58.8}&\textbf{77.4}&\textbf{73.5}&\textbf{86.3}&48.5&\textbf{73.9}&\textbf{76.1}&\textbf{72.9}&\textbf{65.7}&\textbf{62.6}&\textbf{67.5}&\textbf{48.5}&67.2&73.1&68.7&\textbf{13.1}&\textbf{88.3}\\
					\hline
				\end{tabu}
				\label{crf_pascal}
			\end{minipage}
	}}
\end{table*}

Since $(L-\gamma E)$ is a large and sparse matrix which is symmetric and semi-positive definite, we utilize PCG~\cite{pcg} running five times faster than the conventional conjugate method. The whole algorithm presents in algorithm~\ref{alg:1}. Obviously, the coarse annotation expansion procedure for each class has been simplified as matrix decomposition.
\begin{algorithm}[!h]
	\begin{algorithmic}[1]
		\renewcommand{\algorithmicrequire}{\textbf{Input:}}
		\renewcommand{\algorithmicensure}{\textbf{Output:}}
		\REQUIRE Image $I$ and scribble set $S = \{s_k | k = 1,2,...,c\}$;
		\ENSURE confidence maps $A = \{\alpha_k | k = 1,2,..,c\}$;
		\STATE Compute feature vector $X$;
		\STATE Construct affinity matrix $W$ and $D$;
		\STATE Construct Laplacian matrix $L$ and diagonal matrix $E$ with $s$;
		\REPEAT{
			\STATE Update $\alpha_k$ according to Eq.~(\ref{eq:8});
		}
		\UNTIL{$k$ equals to $c$}
		\RETURN{$A$;}
	\end{algorithmic}
	\caption{Algorithm for coarse annotations expansion.}
	\label{alg:1}
\end{algorithm}

\subsection{Noise Control}
We further investigate how to convert the probability distribution map $\alpha$ to a specific label map. Usually by applying a max operation, every pixel will be given a class of maximum probability value, yet we doubt whether this procedure would introduce erroneous labels that negatively lead networks to judge in a wrong way. Thus we derive a strategy to suppress potential noises. For pixel $i$, if $\max(\alpha_{k_i})$, $k=1..c$, is below the threshold, the pixel will be given a label of `ignored'; otherwise it is assigned the label class-$k$, where the value of $\alpha_{k_i}$ is the maximum one among all the classes. It plays a vital role in controlling annotation quality. Detailed discussions and analysis are presented in the experimental section.

\section{Experiments}

To provide a comprehensive evaluation of the proposed method, we structure the experiments as four parts: (1) Quantity analysis in terms of enrichment ratio; (2) Quality analysis in terms of segmentation performance; (3) Efficiency analysis in terms of running time; (4) Ablation Study on feature selection.
\eat{
	Comparison with other annotation refinement methods~(dense-CRF) in terms of enrichment scale (2) Comparison with weakly supervised segmentation models. }

\subsection{Setup}
The experiments are conducted on two real-life datasets, \textit{i.e.,} the Cityscapes~\cite{cityscapes} and PASCAL VOC 2012~\cite{pascalvoc}. The \textbf{Cityscapes} focuses on semantic understanding of urban street, involving 30 classes~(19 classes for evaluation) in $1024\times 2048$ pixel-level semantic labelling. The dataset comprises of 5,000 fine annotations~(2,975 for training, 500 for validation and 1,525 for testing) and 20,000 coarse annotations where 11,900 samples for training and 2,000 for validation (\textit{i.e.,} coarse polygons covering individual objects).  PASCAL VOC 2012 contains 20 object categories and one background category, comprising of 10,582 training data and 1,449 validation data.

Fully-convolutional networks which is based on the VGG 16-layer net~\cite{vgg} are trained for comparison, during which batch size is fixed to 8, and learning rate is fixed to 1e-10 respectively. For evaluation, we use \textit{
	class-wise intersection over union} (IoU) and \textit{pixel-wise accuracy} (Acc.) metrics.

\begin{equation}
		IoU = \frac{TP}{TP + FP +FN},
\end{equation}
where $TP$, $FP$, and $FN$ are the number of true positive, false positive, and false negative pixels, respectively, determined over the whole test set.
\subsection{Quantity Analysis}
In this section, we present the coarse-to-fine enrichment ratio of the proposed method by comparing the generated annotation with the ground-truth dense annotation at the pixel-level. IoU and Acc. are applied to measure the percentage of correctly labelled pixels in two forms. In order to investigate the impact of noise control, two groups of experiments are conducted with a weak threshold~(0.3) as well as a strong threshold~(0.7) respectively. Applying a strong threshold is expected to achieve larger-scale pixel labelling which is beneficial for terrain / road label while it unavoidably hurt the perfection of difficult objects~(\textit{e.g.,} motor bike, bus).

In Table~\ref{crf_cityscapes} and Table~\ref{crf_pascal}, we compare our model with dense-CRF applying on the same coarse annotations and show detailed performances for different classes. For Cityscape dataset, though both CRF and our model boost the performance compared with the coarse annotations~(improving Acc. from 60.9\% to 70.4\% and 82.1\% respectively), there still exists gap between two algorithms. Showing on those `difficult' objects/regions~(\textit{e.g.,} fence, pole), our model shows its superiority on refining boundaries which is cucumber for human annotators, on average rising IoU by 10\%.

\begin{figure*}[!htb]
	\centering
	\subfloat[image]{
		\begin{minipage}[b]{0.18\textwidth}
			\includegraphics[width=1\textwidth]{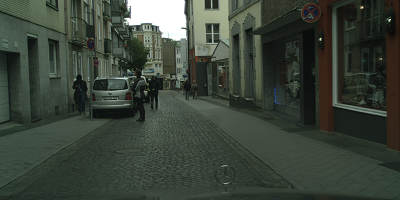}\\
			\vspace{-0.3cm}\\
			\includegraphics[width=1\textwidth]{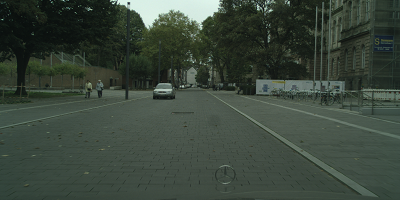}\\
			\vspace{-0.3cm}\\
			\includegraphics[width=1\textwidth]{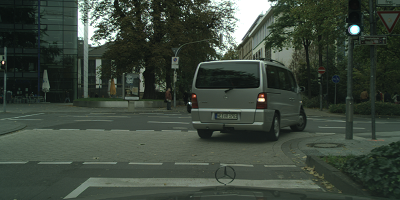}\\
			\vspace{-0.3cm}\\
			\includegraphics[width=1\textwidth]{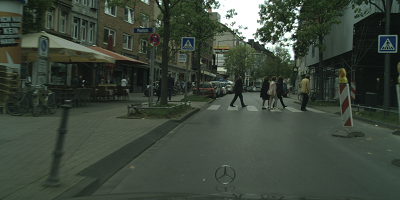}\\
			\vspace{-0.3cm}\\
			\includegraphics[width=1\textwidth]{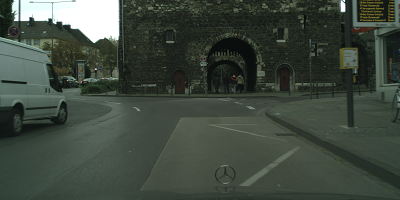}\\
			\vspace{-0.3cm}\\
			\includegraphics[width=1\textwidth]{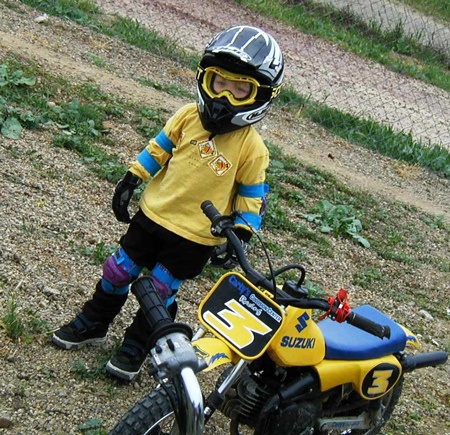}\\
			\vspace{-0.3cm}\\
			\includegraphics[width=1\textwidth]{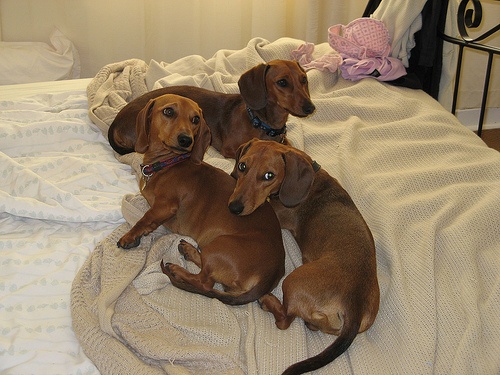}\\
			\vspace{-0.3cm}\\
			\includegraphics[width=1\textwidth]{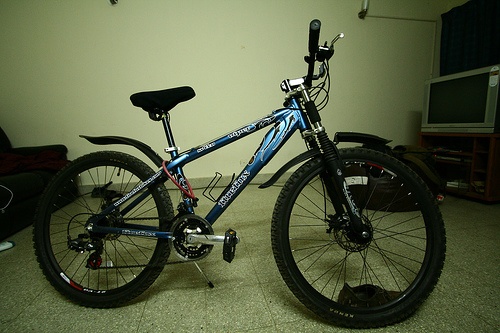}\\
			\vspace{-0.9cm}\\
		\end{minipage}
	}
	\subfloat[GT]{
		\begin{minipage}[b]{0.18\textwidth}
			\includegraphics[width=1\textwidth]{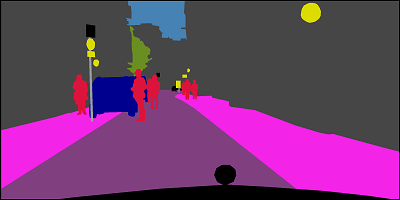}\\
			\vspace{-0.3cm}\\
			\includegraphics[width=1\textwidth]{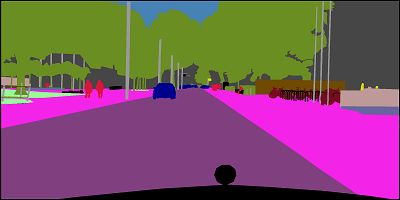}\\
			\vspace{-0.3cm}\\
			\includegraphics[width=1\textwidth]{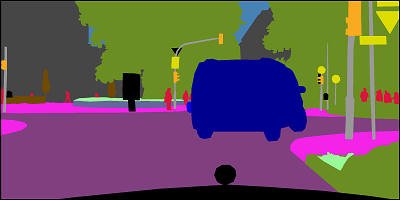}\\
			\vspace{-0.3cm}\\
			\includegraphics[width=1\textwidth]{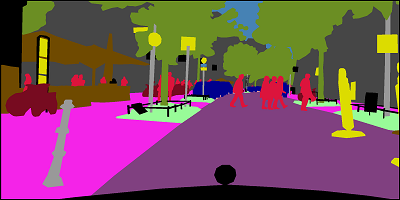}\\
			\vspace{-0.3cm}\\
			\includegraphics[width=1\textwidth]{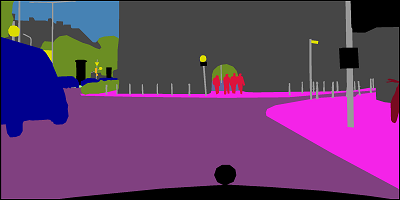}\\
			\vspace{-0.3cm}\\
			\includegraphics[width=1\textwidth]{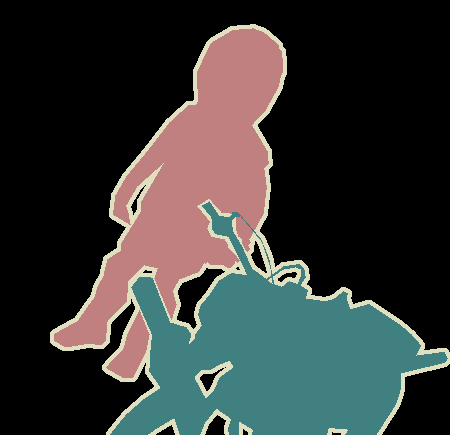}\\
			\vspace{-0.3cm}\\
			\includegraphics[width=1\textwidth]{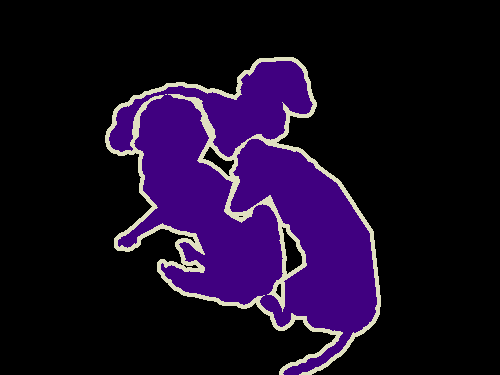}\\
			\vspace{-0.3cm}\\
			\includegraphics[width=1\textwidth]{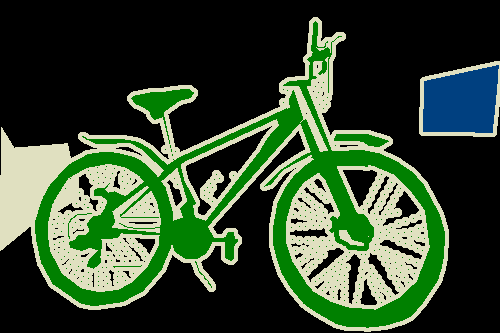}\\
			\vspace{-0.9cm}\\
		\end{minipage}
	}
	\subfloat[Coarse]{
		\begin{minipage}[b]{0.18\textwidth}
			\includegraphics[width=1\textwidth]{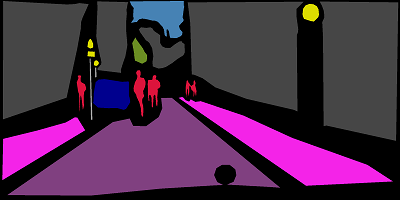}\\
			\vspace{-0.3cm}\\
			\includegraphics[width=1\textwidth]{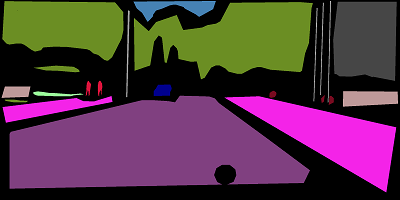}\\
			\vspace{-0.3cm}\\
			\includegraphics[width=1\textwidth]{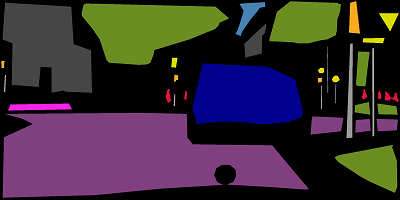}\\
			\vspace{-0.3cm}\\
			\includegraphics[width=1\textwidth]{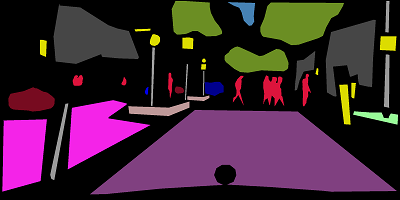}\\
			\vspace{-0.3cm}\\
			\includegraphics[width=1\textwidth]{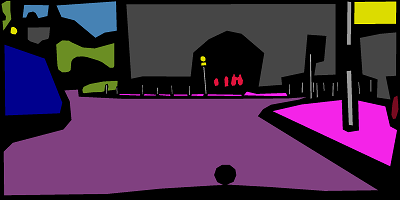}\\
			\vspace{-0.3cm}\\
			\includegraphics[width=1\textwidth]{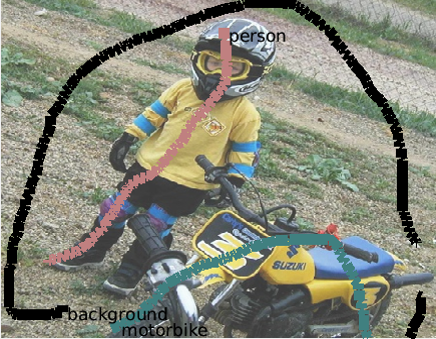}\\
			\vspace{-0.3cm}\\
			\includegraphics[width=1\textwidth]{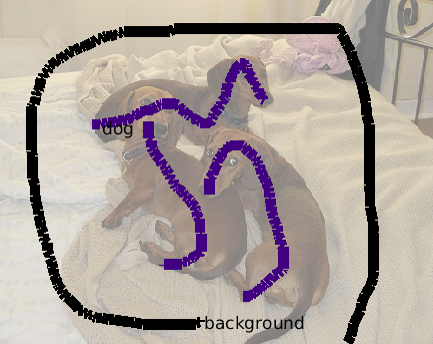}\\
			\vspace{-0.3cm}\\
			\includegraphics[width=1\textwidth]{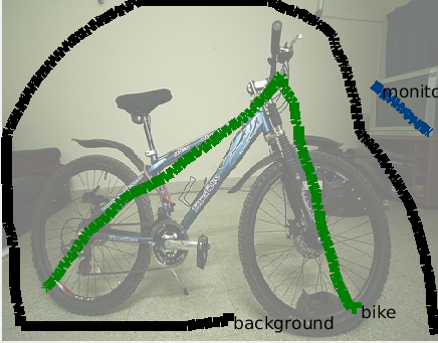}\\
			\vspace{-0.9cm}\\
		\end{minipage}
	}
	\subfloat[denseCRF]{
		\begin{minipage}[b]{0.18\textwidth}
			\includegraphics[width=1\textwidth]{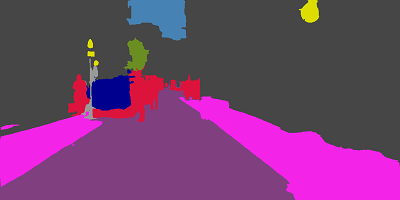}\\
			\vspace{-0.3cm}\\
			\includegraphics[width=1\textwidth]{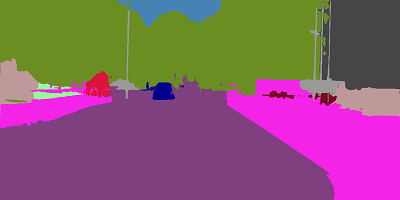}\\
			\vspace{-0.3cm}\\
			\includegraphics[width=1\textwidth]{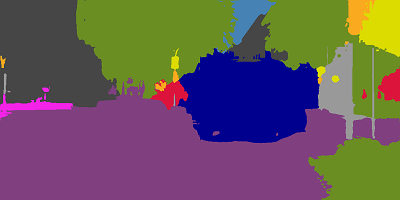}\\
			\vspace{-0.3cm}\\
			\includegraphics[width=1\textwidth]{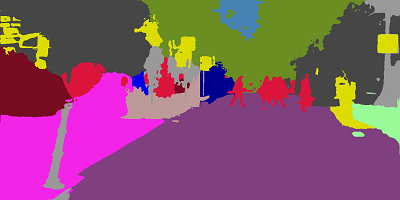}\\
			\vspace{-0.3cm}\\
			\includegraphics[width=1\textwidth]{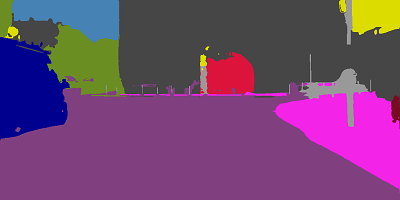}\\
			\vspace{-0.3cm}\\
			\includegraphics[width=1\textwidth]{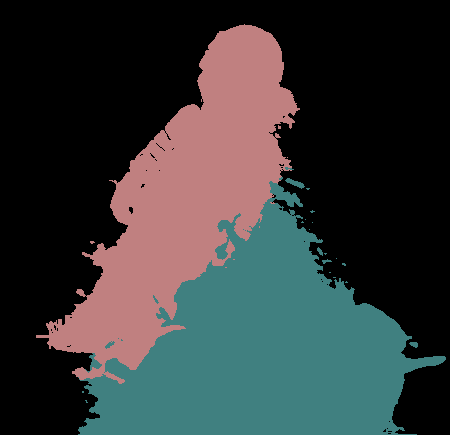}\\
			\vspace{-0.3cm}\\
			\includegraphics[width=1\textwidth]{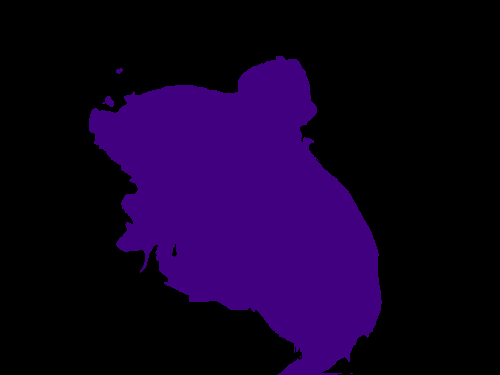}\\
			\vspace{-0.3cm}\\
			\includegraphics[width=1\textwidth]{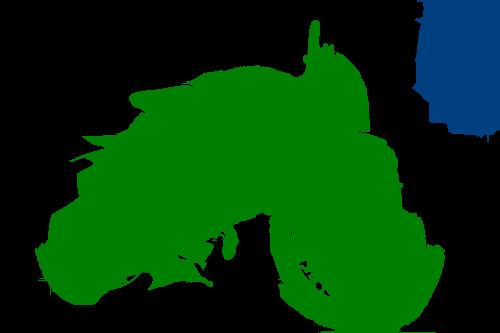}\\
			\vspace{-0.9cm}\\
		\end{minipage}
	}
	\subfloat[ours]{
		\begin{minipage}[b]{0.18\textwidth}
			\includegraphics[width=1\textwidth]{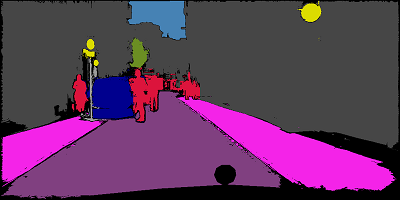}\\
			\vspace{-0.3cm}\\
			\includegraphics[width=1\textwidth]{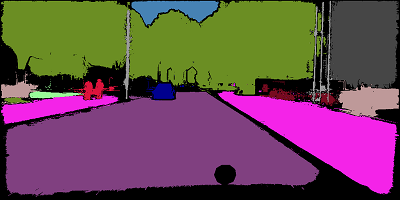}\\
			\vspace{-0.3cm}\\
			\includegraphics[width=1\textwidth]{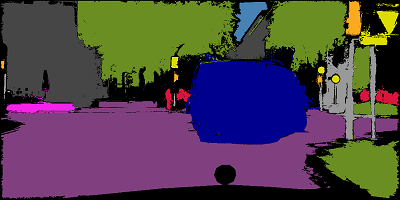}\\
			\vspace{-0.3cm}\\
			\includegraphics[width=1\textwidth]{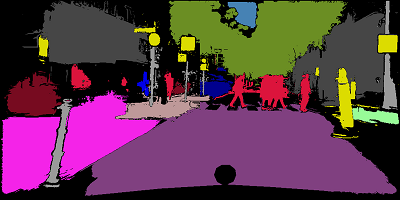}\\
			\vspace{-0.3cm}\\
			\includegraphics[width=1\textwidth]{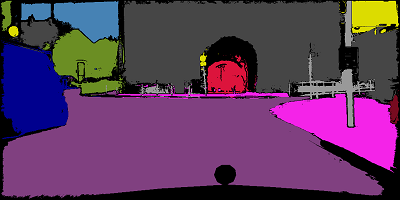}\\
			\vspace{-0.3cm}\\
			\includegraphics[width=1\textwidth]{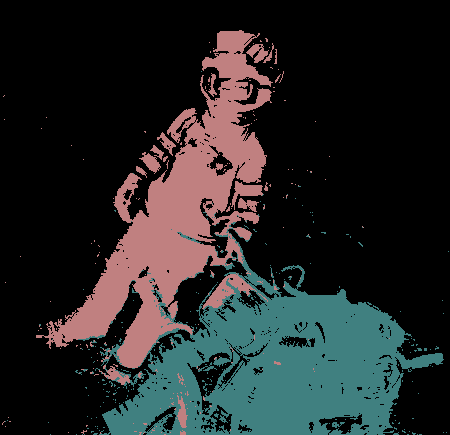}\\
			\vspace{-0.3cm}\\
			\includegraphics[width=1\textwidth]{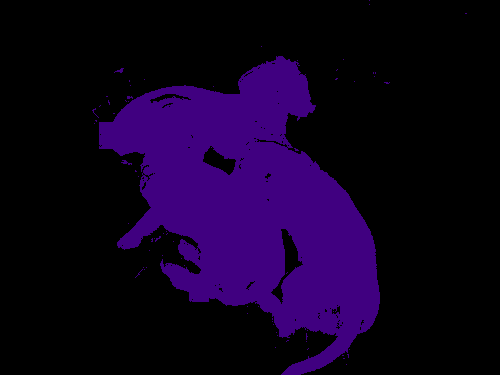}\\
			\vspace{-0.3cm}\\
			\includegraphics[width=1\textwidth]{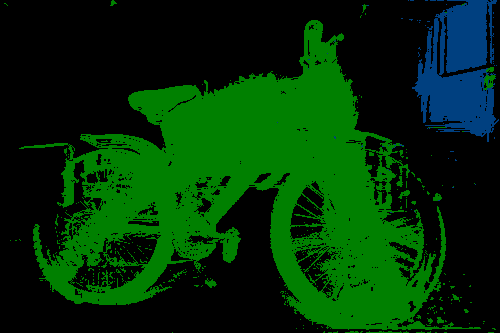}\\
			\vspace{-0.9cm}\\
		\end{minipage}
	}
	\caption{Our results on the Cityscapes (the first five lines) and Pascal VOC 2012 (the last three rows). The coarse annotations we used as raw input are shown in (c). The comparison of refinement quality between dense CRF and our method is shown in (d) and (e).}
	\label{fig:city_samples}
\end{figure*}

Regarding PASCAL VOC 2012 dataset, our model also enhances the positive guidance of coarse annotations, improving mIoU from 25.1\% to 62.9\%~(nearly $4 \times$), meanwhile boosting pixel accuracy from 62.5\% to 88.3\%. Since we treat all unmarked areas as background which also participates in evaluation, scribble gains higher score than what it is expected to achieve. Figure \ref{fig:city_samples} visualizes some real examples of the coarse-to-fine annotations achieved by different methods, from which we can clearly observe the superior performance of our method.

In a brief summary, our method with strict noise control shows its superiority compared with both CRF and that with relaxed noise control. However, there still exists a gap the ground-truth to some extend. We believe the major reason lays on that the coarse annotations we use for expansion ignore many trivial details~(\textit{e.g.}, remote bicycles), which makes the proposed method fail to make up for.

\begin{table}[thb!]
	\centering
	\caption{With the fixed budget, quantitative results of FCN-8s  and annotation time for semantic segmentation on Cityscapes validation set.}
	\label{fix_budget}
	\begin{minipage}[b]{1\hsize}\centering
		\resizebox{1\textwidth}{!}{
			\begin{tabular}{l|l|l|l}
				\hline
				Supervision (fixed budget)              &time(h)  &mIoU(\%) & Acc.(\%) \\
				\hline
				Fine(389 img)           &583  & 32.5     & 57.8     \\
				Coarse(5k img)        &583 & 28.1     & 48.8     \\
				CRF+Coarse(5k img)    &583 & 47.2     &64.6     \\
				Filling(0.7)+Coar.(5k img)&583 & 54.7  &69.1 \\
				Filling(0.3)+Coar.(5k img)&583 &\textbf{55.2}  &\textbf{72.6}
				\\
				\hline
			\end{tabular}
		}
	\end{minipage}
\end{table}

\begin{table}[thb!]
	\centering
	\caption{With unlimited budget, quantitative results of FCN-8s  and annotation time for semantic segmentation on Cityscapes validation set.}
	\label{unlimit_budget}
	\begin{minipage}[b]{1\hsize}\centering
		\resizebox{1\textwidth}{!}{
			\begin{tabular}{l|l|l|l}
				\hline
				Supervision (unlimited budget)                 &time(h) & mIoU(\%) &Acc.(\%) \\
				\hline
				Fine(2,975 img)            &4,463  &58.1     & 77.8     \\
				Coarse(2,975 img)          &347   &19.6     &29.5    \\
				Fil.+Coar.(2,975 img)      &347  &49.7      &68.4 \\
				Coarse(11.9k img)         &1,388  &56.2     & 74.9     \\
				CRF+Coarse(11.9k img)     &1,388  & 65.7     &81.6     \\
				Fil.(0.3)+Coar.(11.9k img) &1,388 & 68.1  &82.3 \\
				Fil.(0.7)+Coar.(11.9k img) &1,388 & \textbf{69.4}  &\textbf{82.9}
				\\
				\hline
			\end{tabular}
		}
	\end{minipage}
\end{table}
\subsection{Quality Analysis}
In this work, the quality of the enriched annotation is evaluated by feeding the generated annotations into supervised segmentation models and comparing the segmentation performance with the ground-truth. We consider two scenarios for performance study, where one is segmentation with fixed annotation budget and the other one is segmentation with unlimited annotation budget. The reason of considering annotation budget into the training step is that the demand of high quality dense annotated data is extremely labor- and time- consuming. As \cite{cityscapes} states, fine annotations with quality control require more than 1.5 hours on average for a single image while coarse annotations only take 7 minutes. Please note that, compared to the time cost of manual annotating, the computational cost of the proposed model generating enriched annotations based on the original coarse annotations are negligible.

\subsubsection{With Fixed Annotation Budget}
Given a fixed annotation time budget, what is the right strategy to obtain the best segmentation model possible? We investigate the problem with the Cityscape dataset by fixing the annotation time to be $7min\times 5,000 = 35,000 min$, that it would take to annotate all the 5,000 images with coarse annotations. For fine annotation supervision, we then compute the numbers of images $N = 35,000min / 90min \approx 389$ that are possible to be labelled within the same amount of time. A number of $N$ images are randomly sampled from the training dataset, with which a FCN-8s segmentation model is trained from scratch.

\begin{figure}[!htb]
	\centering
	\includegraphics[width=1\linewidth]{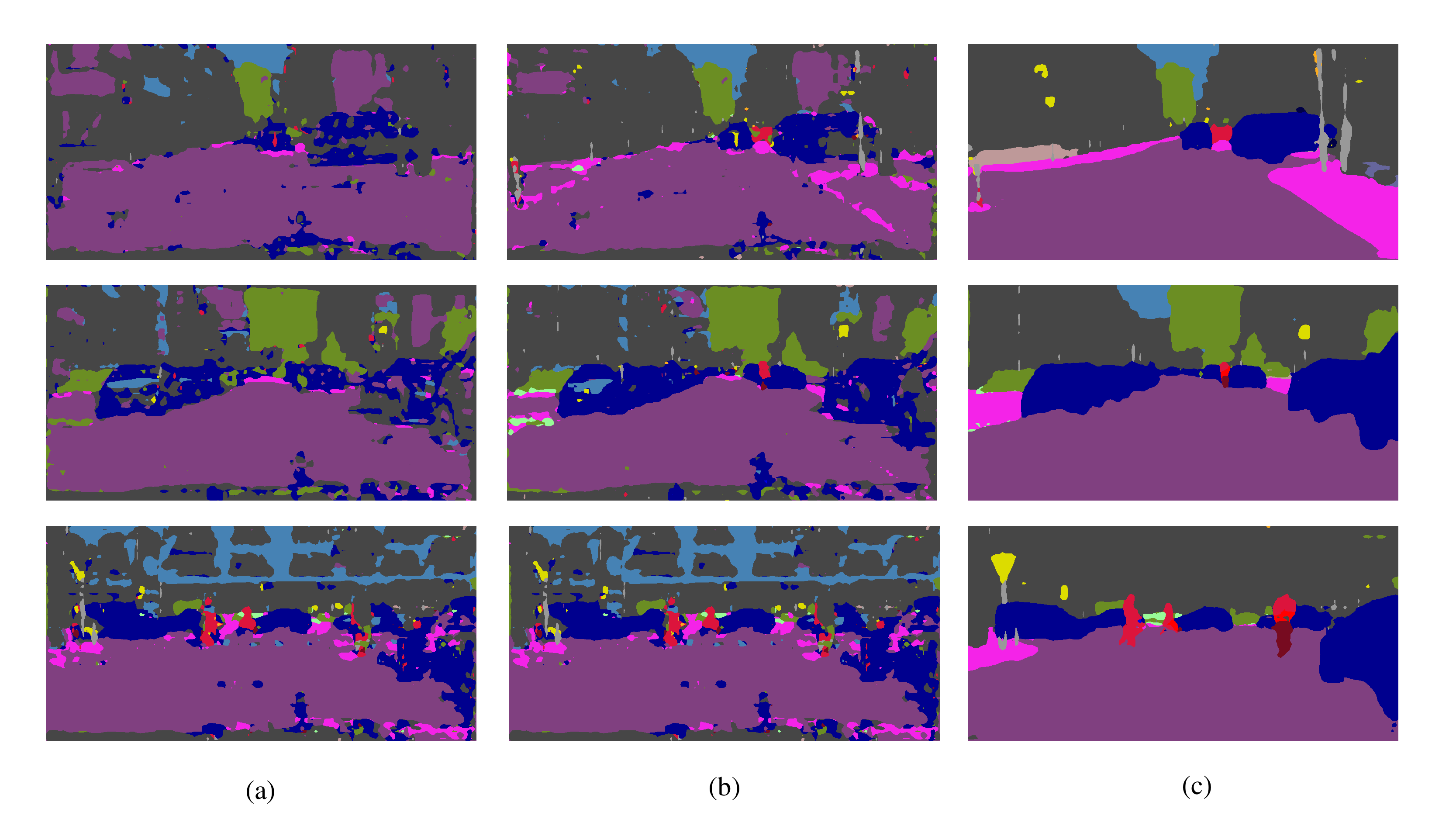}
	\caption{Qualitative examples of FCN-8s trained with (a) 5,000 coarse annotations, (b) 389 fine annotations, (c) 5,000 enriched annotations by the proposed model.}
	\label{fig:fcn_cityscapes}
\end{figure}

Table~\ref{fix_budget} reports the resulting accuracy of fully-supervised~(32.5\% mIoU), supervised with coarse annotations~(28.1\% mIoU), with CRF processed annotations~(47.2\% mIoU) and with the proposed enriched annotations~(55.2\% mIoU) respectively. Since the training model is far from overfitting~(with tiny training dataset), we observe that our proposed model with a weak threshold performs slightly better than that with a strong threshold. Some examples of the training performance are visualised in Figure \ref{fig:fcn_cityscapes}.

\subsubsection{With Unlimited Annotation Budget }
We also compare both annotation time and quality of our enriched annotations with strongly-supervised annotations with much larger annotation budgets, as a reference for what might be achieved by our method if given more resources. As we could see in Table~\ref{unlimit_budget}, a FCN trained with fully-supervised annotations that requires 4,463 hours to annotate reports 58.1\% mIoU, while that with coarse annotations consuming only 1,388 hours could reach 65.7\% mIoU~(CRF processed), or even 69.4\% mIoU~(ours). We could see,
 as the training data size goes up, our model with a strict threshold outperforms, strongly proving validity of our noise control. In addition, the FCN trained with 2,975 enriched annotations~(49.7\%) has a slight degradation in comparison with that with a 5,000 dense annotations~(58.1\%), yet it is obtained at a $13\times$ cheaper cost.
\subsubsection{Comparison with Weakly-supervised Methods}

Without regard to time budget, we compare the fully-supervised segmentation model trained with the enriched annotations generated by our model with the state-of-the-art weakly-supervised methods. This set of experiments are conducted on the PASCAL VOC 2012 as it provides each image various coarse annotations for the usage of different weakly-supervised models. Our annotation enrichment is performed on the coarse annotations of ``scribble". ``Scribble" is a very economical coarse annotation, which averagely takes only 25 seconds for human to label one image with 20 object categories, according to \cite{scribbleup, point,extreme}. However, it is more challenging to expand compared to generating enriched annotations from coarse polygons.

\eat{setting, we take more challenging user scribbles as coarse annotations to verify generalization of the proposed method.
	According to the record~\cite{point,extreme}, it averagely takes an annotator 25 seconds to label an image with 20 object categories, while a box annotation, a point, a image tag takes 59 seconds, 22 seconds, 19 seconds on average respectively.
}
\begin{table}[htb!]
	\centering
	\caption{Comparisons of weakly-supervised methods on the PASCAL VOC 2012 validation set, using different forms of annotations.}
	\resizebox{.49\textwidth}{!}{
		\begin{tabular}{l|l|l|l}
			\hline
			Supervision     &time(h) &annotations     & mIoU(\%)  \\
			\hline
			Grabcut\cite{rother2004grabcut}+FCN       &73.5 &scribble      & 49.1        \\
			LazySnap.\cite{lazysnapping}+FCN  &73.5&scribble      & 53.8         \\
			CRF+FCN               &73.5&scribble       & 57.7 \\
			Scribblesup\cite{scribbleup} &73.5 &scribble &63.1 \\
			Fil.(0.7)+FCN  &73.5&scribble      & 65.2  \\
			Fil.(0.3)+FCN  &73.5&scribble  & \textbf{67.7} \\
			\hline
			Point\cite{point}  &64.7 &point        &42.7\\
			\hline
			BoxSup\cite{boxsup}&173.4 &box            &62.0\\
			WSSL\cite{wssl}    &173.4 &box          &60.6\\
			\hline
			MIL-FCN\cite{image}     &55.8 &image      &25.1\\
			AE-PSL\cite{AE-PSL}     &55.8 &image      &47.8\\
			AF-SS\cite{AF-SS}     &55.8 &image      &52.6\\
			\hline
		\end{tabular}
	}
	\label{unsupervised}
\end{table}

Table~\ref{unsupervised} shows the segmentation results from our model and the state-of-the-art weakly-supervised methods using different ways of annotations. Note that the outcome from Interactive segmentation~(\textit{i.e., Grabcut and LazySnapping}) and refinement methods~(CRF and ours) is used for FCN training, then yielding comparable results. They image-level annotations are the most economical among others, they have limited performance (\textit{e.g.,} 25.1\% mIoU) or draw support from other localization prior (\textit{e.g.,} AF-SS). The point-based method reaches 42.7\% mIoU, yet not competitive with Box-supervised approaches~(\textit{e.g.,} BoxSup 62.0\% mIoU). Though, scribble method achieves higher score~(\textit{e.g., ours 67.7\%}) with light work load~(73.5 hours) compared with box-based methods~(173.4 hours).

In groups of scribble-based approaches, our method with strict quality control performs the best compared with the rest. Scribblesup~\cite{scribbleup} achieves a comparable result~(63.1\%) with ours. However, it is jointly learned with FCN for \emph{multiple} iterations with extra information input for fine-tuning, which could take much longer time for training. In contrast, our trained models only experience once training procedure.

\subsection{Efficiency Analysis}
In this section, we mainly investigate the running time of each algorithm, i.e., Dense CRF and the proposed method. The experiments have been conducted on a server with Intel Xeon(R) CPU E5-2660 and two Tesla K40c GPU cards. The implementation of CRF is in Python and the implementation of our method is based on MATLAB mixed with C++. We sample 20 images from each datasets and calculate the average running time as shown in Table \ref{time}. The size of images from Cityscapes is $1024\times512$ and that from PASCAL VOC is $500\times375$. What we can clearly observe is that our algorithm performs approximately $1.9\times$ faster compared with dense CRF under the same setting, which makes the proposed method more practicable and suitable for realistic application. 
\begin{table}[htb!]
	\centering
	\caption{Comparison on running time of dense CRF and the proposed method.}
	\resizebox{.49\textwidth}{!}{
		\begin{tabular}{l|c|c|c}
			\hline
			Method     &Number of Classes &Cityscapes    &PASCAL VOC  \\
			\hline
			CRF      &10 &2.37s      &1.19s         \\
			Fil(0.7)  	&2	&1.03s  &0.37s\\
			\hline
		\end{tabular}
	}
	\label{time}
\end{table}

\subsection{Ablation Study on feature selection}
\begin{figure*}[htb!]
	\centering
	\subfloat[image]{
		\begin{minipage}[b]{0.25\textwidth}
			\includegraphics[width=1\textwidth]{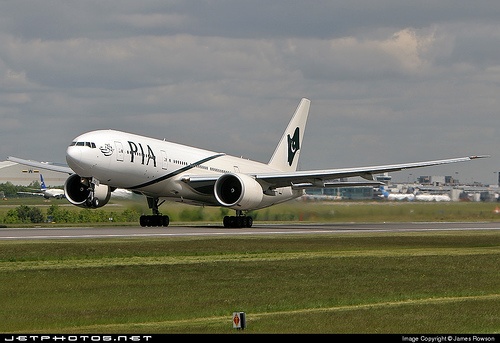}\\
			\vspace{-0.3cm}\\
			\includegraphics[width=1\textwidth]{images/2007_000783.jpg}\\
			\vspace{-0.3cm}\\
			\includegraphics[width=1\textwidth]{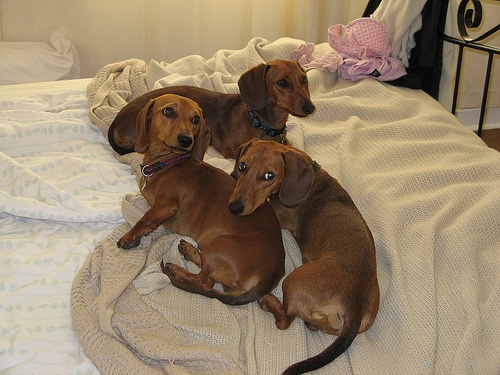}\\
			\vspace{-0.3cm}\\
			\includegraphics[width=1\textwidth]{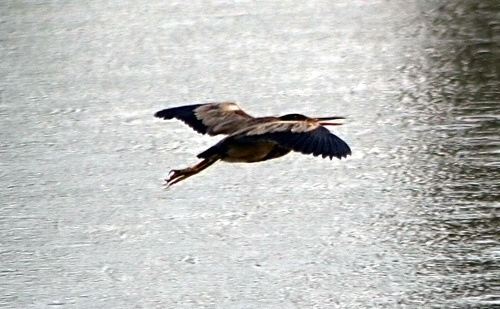}\\
			\vspace{-0.3cm}\\
			
		\end{minipage}
	}
	\subfloat[ground-truth]{
		\begin{minipage}[b]{0.25\textwidth}
			\includegraphics[width=1\textwidth]{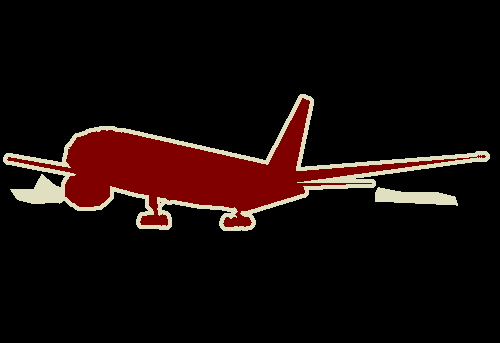}\\
			\vspace{-0.3cm}\\
			\includegraphics[width=1\textwidth]{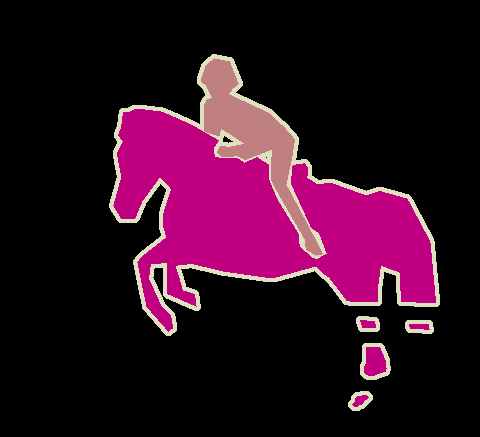}\\
			\vspace{-0.3cm}\\
			\includegraphics[width=1\textwidth]{images/gt_1239.png}\\
			\vspace{-0.3cm}\\
			\includegraphics[width=1\textwidth]{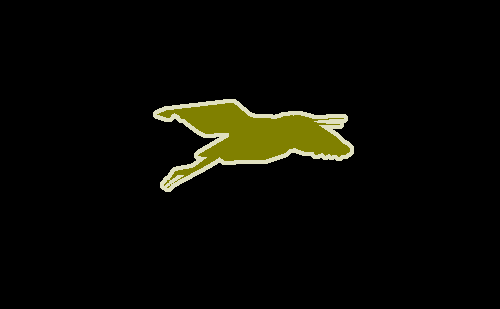}\\
			\vspace{-0.3cm}\\
			
		\end{minipage}
	}
	\subfloat[Local]{
		\begin{minipage}[b]{0.25\textwidth}
			\includegraphics[width=1\textwidth]{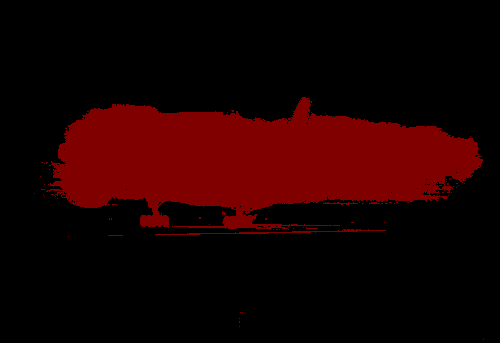}\\
			\vspace{-0.3cm}\\
			\includegraphics[width=1\textwidth]{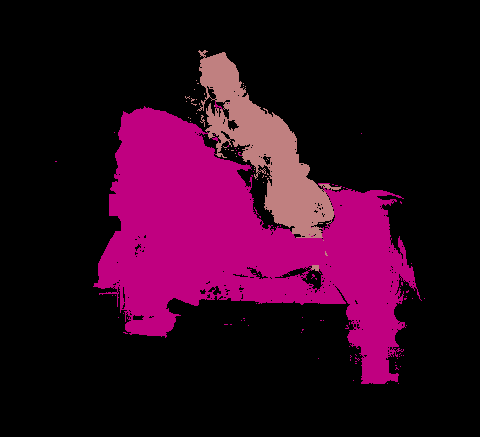}\\
			\vspace{-0.3cm}\\
			\includegraphics[width=1\textwidth]{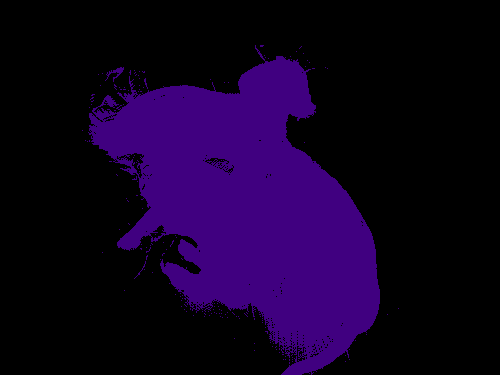}\\
			\vspace{-0.3cm}\\
			\includegraphics[width=1\textwidth]{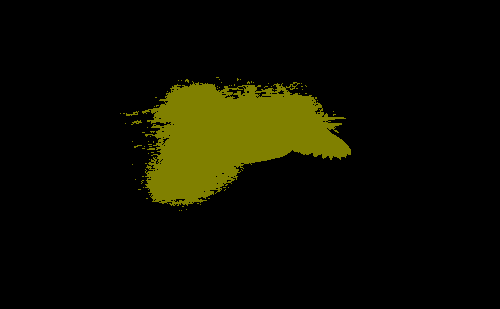}\\
			\vspace{-0.3cm}\\
			
		\end{minipage}
	}
	\subfloat[+Ftmap]{
		\begin{minipage}[b]{0.25\textwidth}
			\includegraphics[width=1\textwidth]{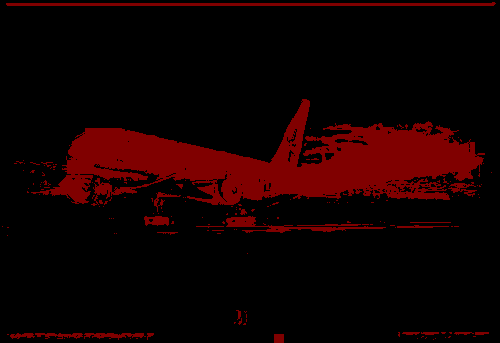}\\
			\vspace{-0.3cm}\\
			\includegraphics[width=1\textwidth]{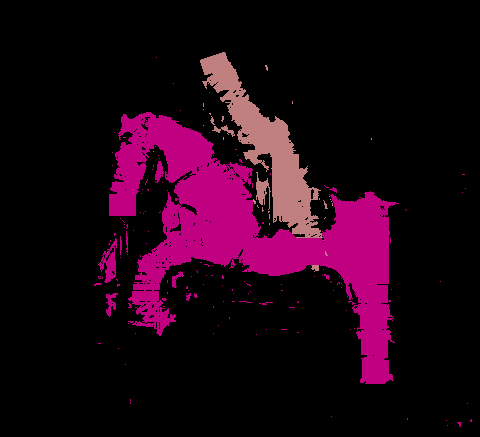}\\
			\vspace{-0.3cm}\\
			\includegraphics[width=1\textwidth]{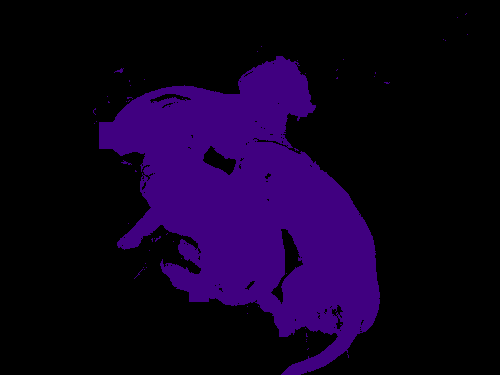}\\
			\vspace{-0.3cm}\\
			\includegraphics[width=1\textwidth]{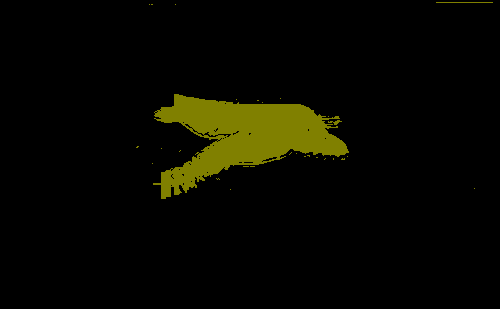}\\
			\vspace{-0.3cm}\\
			
		\end{minipage}
	}
	\caption{Qualitative results of annotation enrichment algorithm (c)~with pure RGB+coordinates features and (d)with additional feature maps on the PASCAL VOC 2012 dataset. Best viewed in electronic version.}
	\label{fig:color}
\end{figure*}

In this section, we mainly discuss the validity of our feature space construction which is introduced in Section 3.5. In terms of quantitative results,  we report the IoU per class and pixel-wise accuracy on PASCAL VOC 2012 dataset in comparison of the proposed methods with original feature space (including RGB and coordinates information) and with additional feature maps in Table \ref{color}. With the same setting, the proposed method that incorporates regional patterns gain extra $1.4\%$ accuracy on annotation enrichment. Beyond quantitative results, we also give the visualizations of refined annotation with different feature combination. From the Figure \ref{fig:color} we could see that, the rightest column of enriched labeled data show more accurately on boundary prediction and capture more connection between pixels. Those results suggest that our feature space constructed with robust feature maps have a positive impact on the annotation enrichment of labeled data. 
\begin{table*}
	\caption{Comparisons of annotation enrichment algorithm with pure local (e.g., RGB, coordinates) features and with additional feature maps (\textit{the last line}) on the PASCAL VOC 2012 dataset \textit{w.r.t} IoU and Accuracy.}
	\resizebox{1\linewidth}{!}{
		\begin{tabu}to 1.06\textwidth{X[3,c]|X[c]X[1.3,c]X[c]X[c]X[1.1,c]X[c]X[c]X[c]X[c]X[1.3,c]X[1,c]X[1.2c]X[c]X[c]X[c]X[c]X[c]X[c]X[1.3,c]X[1,c]X[1.3,c]|X[1.5,c]}
			\hline
			Method & bg & acr & bic& bir & boat &bot & bus & car & cat & chair & cow & din & dog & hor & mot & pcr & pot & shc & sof & tra & tv &acc.\\
			\hline
			Coarse &60.3&34.3&17.6&26.1&31.2&27.6& 19.3&22.8&20.0&27.5&25.2&22.3&21.3&26.2&22.4&25.8&24.2&25.6&23.4&20.8&2.7&62.5\\
			CRF &83.1&48.1&22.0&51.6&46.0&58.5& 68.6&61.1&74.1&47.3&63.9&75.8&68.5&62.1&58.9&61.9&47.2&61.3&74.0&67.9&9.8&83.6\\
			Local &86.3&56.2&31.1&65.1&50.4&55.6& 78.8&70.5&84.1&\textbf{55.0}&71.1&75.4&71.5&58.5&65.7&66.4&40.9&\textbf{70.3}&73.1&67.9&11.2&86.9\\
			+Ftmap &\textbf{87.0}&\textbf{58.8}&\textbf{33.2}&\textbf{76.6}&\textbf{51.9}&\textbf{58.8}&\textbf{77.4}&\textbf{73.5}&\textbf{86.3}&48.5&\textbf{73.9}&\textbf{76.1}&\textbf{72.9}&\textbf{65.7}&\textbf{62.6}&\textbf{67.5}&\textbf{48.5}&67.2&\textbf{73.1}&\textbf{68.7}&\textbf{13.1}&\textbf{88.3}\\
			\hline
		\end{tabu}
	}
	\label{color}
\end{table*}

\section{Conclusion and Future Work}
We have presented a generalized annotation enrichment algorithm that aims to extend coarse annotations to a finer scale, alleviating the density and efficiency problem of annotation. Experiment results show it is also applicable for refining challenging scribbles and the trained model outperforms among other state-of-the-art weakly-supervised segmentation models. The proposed method with strict noise control is viable for suppressing errors and thereby enhancing positive guidance for training semantic segmentation models. Since our method is agnostic to fully-supervised segmentation models , we plan to train more complex and powerful neural networks (\textit{e.g.,} DeepLabv3, ResNet) with our enriched annotations, expected to see a better performance on existing datasets.

\section{Acknowledgement}
This work is partially supported by ARC FT130101530 and NSFC No. 61628206.

\newpage

\bibliographystyle{ACM-Reference-Format}
\bibliography{sample-sigconf}

\end{document}